\documentclass[conference]{IEEEtran}
\IEEEoverridecommandlockouts

\usepackage{pifont} 

\usepackage{siunitx}
\usepackage{multirow}
\usepackage{amsmath}
\usepackage{makecell}
\usepackage{booktabs}
\usepackage{array}
\usepackage{booktabs} 
\usepackage{graphicx}
\usepackage{csquotes}
\usepackage{subcaption}
\usepackage{paralist}
\usepackage{comment}
\usepackage[table,xcdraw]{xcolor}
\usepackage{tabularx}
\usepackage{tikz}

\usetikzlibrary{arrows.meta, positioning, shapes, fit, calc}

\usepackage[ruled,vlined,linesnumbered]{algorithm2e}
\usepackage{amssymb}

\usepackage[style=ieee]{biblatex}
\usepackage[section,numberedsection=autolabel]{glossaries}
\makeglossaries

\newglossaryentry{function}{name=function,%
    description={A single, self-contained piece of software, that performs a certain function}}

\newglossaryentry{feature}{name=feature,
    description={Composed of one or more functions connected together using a certain runtime environment, usually corresponds to a certain use-case}}

\newglossaryentry{runtime environment}{name=runtime environment,
    description={Communication middleware and virtualization mechanisms}}

\newglossaryentry{ECU}{name=ECU,
    description={Electronic Control Unit. It is an electronic device in a vehicle that is responsible for a single function}}

\newglossaryentry{TDD}{name=TDD,
    description={Test-Driven Development is a software development methodology that centers on the iterative creation of unit tests prior to the implementation of functional code~\cite{ref40:Beck2002} This approach mandates that a test case specifying the desired behavior of a code unit be written before the production code itself. As development progresses, the test suite continuously executes. New code is only written if it fulfills the requirements outlined in a failing test}}

\newglossaryentry{FDD}{name=FDD,
    description={Feature-Driven Development. It is a paradigm where the software system is iteratively developed in a series of steps, starting with an abstract model of the system, followed by extraction of a set of desired features, and ending with feature implementation and integration~\cite{ref41:Palmer2001}}}

\newglossaryentry{MBSE}{name=MBSE,
    description={Model-Based Systems Engineering is a formalized methodology within systems engineering that emphasizes using models as the primary means of information exchange and system representation~\cite{ref42:Incose2023}. This contrasts with traditional document-centric approaches. MBSE centers on creating and leveraging domain-specific models or metamodels, which capture system requirements, design, analysis, and verification elements throughout the development lifecycle}}

\newglossaryentry{contract}{name=contract,
    description={Design by contract is a software development methodology that emphasizes the explicit definition of formal contracts between software components~\cite{ref43:Mitchell2002}. These contracts specify preconditions (what must be true before a component is used), postconditions (what must be true after execution), and invariants (conditions that must always hold true). Design by contract can be enforced through runtime assertions, unit tests, or even integrated into a programming language's syntax. This approach enhances software reliability, eases debugging, and facilitates code comprehension}}

\newglossaryentry{metamodel}{name=Metamodel,
    description={Defines the language of system description by specifying abstract entities that are part of the system, a set of possible relations between them, and their attributes}}

\newglossaryentry{instance model}{name=Instance model,
    description={A model generated from the given metamodel, populated with actual objects with concrete attribute values; an implementation of the system described in the language of the metamodel}}

\newglossaryentry{OMG}{name=OMG,
    description={Object Management Group}}

\newglossaryentry{LLM}{name=LLM,
    description={Large Language Model}}

\newglossaryentry{OCL}{name=OCL,
    description={Object Constraint Language}}

\newglossaryentry{RACE}{name=RACE,
    description={Centralized Platform Computer Based Architecture for Automotive Applications}}

\newglossaryentry{Ecore}{name=Ecore,
    description={Language of the metamodel used in Eclipse Modeling Framework}}

\newglossaryentry{OEM}{name=OEM,
    description={Original Equipment Manufacturer}}

\makeatletter 
\newcommand{\linebreakand}{%
  \end{@IEEEauthorhalign}
  \hfill\mbox{}\par
  \mbox{}\hfill\begin{@IEEEauthorhalign}
}
\makeatother 

\begin{document}

\title{PRAM-R: A Perception-Reasoning-Action-Memory Framework with LLM-Guided Modality Routing for Adaptive Autonomous Driving
}



\author{
\IEEEauthorblockN{Yi Zhang, Xian Zhang, Saisi Zhao, Yinglei Song, Chengdong Wu, Nenad Petrovic and Alois Knoll}
\IEEEauthorblockA{\textit{Chair of Robotics, Artificial Intelligence and Embedded Systems} \\
\textit{Technical University of Munich (TUM)}\\
Munich, Germany \\
\{yi1228.zhang, xian.zhang, saisi.zhao, yinglei.song, chengdong.wu, nenad.petrovic, k\}@tum.de}
}

\maketitle

\begin{abstract}
Multimodal perception enables robust autonomous driving but incurs unnecessary computational cost when all sensors remain active. This paper presents PRAM-R, a unified Perception–Reasoning–Action–Memory framework with LLM-Guided Modality Routing for adaptive autonomous driving. PRAM-R adopts an asynchronous dual-loop design: a fast reactive loop for perception and control, and a slow deliberative loop for reasoning-driven modality selection and memory updates. An LLM router selects and weights modalities using environmental context and sensor diagnostics, while a hierarchical memory module preserves temporal consistency and supports long-term adaptation. We conduct a two-stage evaluation: (1) synthetic stress tests for stability analysis and (2) real-world validation on the nuScenes dataset. Synthetic stress tests confirm 87.2\% reduction in routing oscillations via hysteresis-based stabilization. Real-world validation on nuScenes shows 6.22\% modality reduction with 20\% memory recall while maintaining comparable trajectory accuracy to full-modality baselines in complex urban scenarios. Our work demonstrates that LLM-augmented architectures with hierarchical memory achieve efficient, adaptive multimodal perception in autonomous driving. 
\end{abstract}


\section{Introduction}
\begin{figure*}[t]
    \centering
    \includegraphics[width=0.93\textwidth]{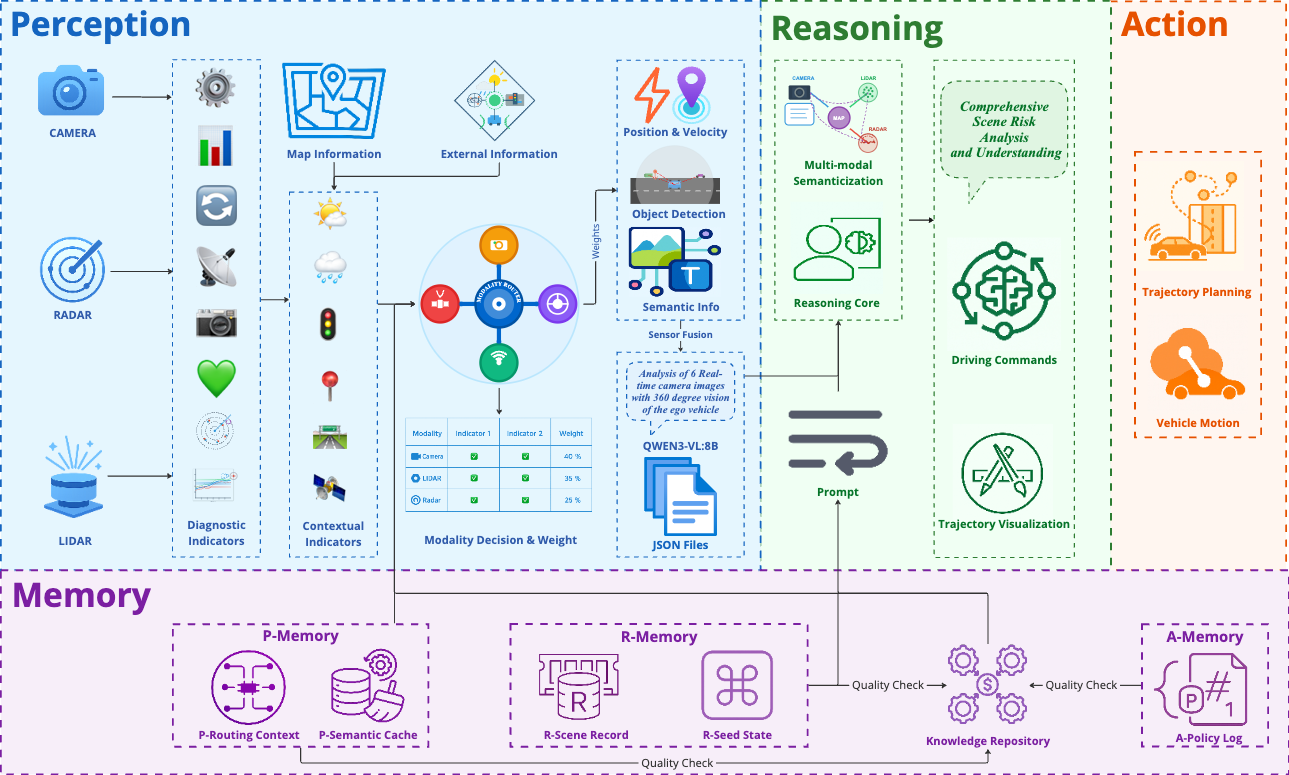} 
    \caption{Perception-Reasoning-Action-Memory Framework with modality routing for complex scene interpretation and motion control in autonomous driving: (1) Perception-layer, (2) Language-layer, (3) Action-layer, and (4) Memory-Layer. }
    \label{fig:PRAM-R}
\end{figure*}

Autonomous driving relies on multiple modalities (camera, LiDAR, radar and map) to perceive complex environments. While multimodal fusion enhances generalization and robustness in perception~\cite{huang2021makes, xiao2020multimodal}, it introduces efficiency challenges: continuously activating all modalities incurs unnecessary computational overhead and latency, with diminishing accuracy gains when redundant or low-quality modalities are included~\cite{liu2018efficient}.

Activating all sensors simultaneously is impractical under limited computing resources and real-time constraints. Different scenarios require distinct modality combinations. For example, cameras suffice for highway lane keeping in clear weather, while urban intersections at night or in rain benefit from radar. Thus, adaptively selecting the most informative and reliable modalities is essential for maintaining perception robustness and optimizing computation, energy, and sensor longevity. These factors motivate an intelligent modality routing mechanism that dynamically activates sensors according to scene context and system state. Yet, existing approaches rely on heuristic rules or attention mechanisms that lack high-level reasoning capabilities, and most fusion frameworks either employ fixed modality selection strategies or are tailored to specific driving scenarios~\cite{prakash2021multi, he2024efficient, khalil2022exploiting}, making them difficult to generalize and often unable to balance perception accuracy and computational efficiency. 
Recent advances in large language models (LLMs) have enabled LLM-guided perception~\cite{sathyam2025foundation}, extending even to modality routing. However, applying LLMs to real-time driving faces two key challenges: First, their sequential, computation-heavy reasoning introduces latency when invoked per frame; Second, the absence of memory prevents reusing past routing experiences, leading to redundant reasoning and poor long-term adaptation.

To address these challenges, following the framework described in~\cite{zhang2025unified}, we propose \textbf{PRAM-R}, a \textit{Perception–Reasoning–Action–Memory} framework with \textit{LLM-Guided Modality Routing} for adaptive autonomous driving. 
Specifically, large language model serves as a \textit{modality router}, selecting and weighting modalities based on scene context and sensor health. 
Our contributions are summarized as follows:

\begin{itemize}
    \item We propose \textbf{PRAM-R}, a novel Perception-Reasoning-Action-Memory framework that integrates LLM-guided modality routing into the autonomous driving pipeline for adaptive and efficient multimodal perception.   
    \item We introduce a \textbf{reasoning-driven routing mechanism} that dynamically selects and weights sensor modalities through integrated analysis of sensor diagnostics and environmental context, optimizing the trade-off between computational efficiency and perceptual reliability.
    \item We design an \textbf{asynchronous dual-loop architecture} decoupling fast reactive control from deliberative reasoning adaptation, achieving low-latency real-time operation while maintaining robust performance across diverse driving scenarios.
    \item We develop a \textbf{hierarchical memory module} that caches routing decisions to reduce inference overhead, enforces temporal consistency across frames, tracks evolving sensor reliability patterns, and preserves contextual knowledge for continual environmental adaptation.
\end{itemize}

\section{Related Work}
\label{sec:related_work}

\subsection{Multimodal Perception and Fusion}
Multimodal fusion has been widely explored for autonomous driving, combining cameras, LiDAR, radar, and HD maps for robust perception. 
Early fusion methods integrate features at the input or BEV level, while late fusion aggregates detection results from individual modalities.
Representative architectures include BEVFusion~\cite{liu2022bevfusion}, TransFuser~\cite{chitta2022transfuser}, and MVFNet~\cite{wu2021mvfnet}, which achieve high detection accuracy by leveraging complementary sensing modalities. 
However, these approaches assume that all sensors are always reliable and active, leading to redundant computation and degraded performance under sensor failures or adverse conditions.

\subsection{Modality Routing and Sensor Diagnostics}
Recent research has shifted from static fusion to adaptive, reliability-aware routing that dynamically selects or weights sensor modalities. Methods such as AdaFusion~\cite{lai2022adafusion} and DynMM~\cite{xue2023dynamic} learn modality gating through attention or uncertainty cues, while SamFusion~\cite{palladin2024samfusion} adapts fusion weights to adverse weather. In parallel, quantitative diagnostics have been developed to assess sensor health using LiDAR density and echo ratio~\cite{heinzler2020cnn}, radar cross section~\cite{sevgi2013rcs}, and vision indicators like brightness and contrast~\cite{sakaridis2019gcma}. These studies provide useful priors for evaluating reliability but largely rely on heuristics and lack reasoning-driven decision integration.

\subsection{LLM and Memory-Augmented Driving Agents}
Large language and vision-language models have recently been explored for driving-scene reasoning and decision making. 
Works such as DriveGPT4~\cite{xu2024drivegpt4} and OmniDrive~\cite{wang2024omnidrive} employ generative reasoning to describe scenes and propose high-level actions.
Meanwhile, temporal-memory agents such as DriveDreamer~\cite{wang2024drivedreamer}, DriveLM~\cite{sima2024drivelm}, and Wayformer~\cite{nayakanti2022wayformer} capture long-term spatio-temporal dependencies for trajectory prediction and planning.
These advances highlight the potential of integrating language-based reasoning and memory, yet current systems operate on fixed sensor inputs and do not adapt sensing configurations according to environmental or diagnostic feedback.


However, none of the above approaches jointly integrate LLM-driven modality routing with hierarchical memory for long-term adaptation, which motivates our PRAM-R.

\section{Methodology}
\label{sec:method}
\subsection{Framework Overview}

To enable adaptive autonomous driving  and address the challenges of redundant computation as well as inconsistent multimodal reliability under diverse and uncertain conditions, we propose \textbf{PRAM-R}, a unified \textbf{Perception–Reasoning–Action–Memory} framework with \textbf{LLM-Guided (Qwen3-VL-8B) Modality Routing}. As shown in Fig.~\ref{fig:PRAM-R}, the architecture integrates four tightly coupled modules (Perception, Reasoning, Action, and Memory) with an LLM-based Modality Router embedded in the perception layer. The reasoning and action layers retain the designs in~\cite{zhang2025unified}, while the perception layer is enhanced with LLM-guided modality routing and a new fusion strategy for adaptive multimodal perception. In addition, a Memory module is newly introduced to enable contextual adaptation and maintain long-term consistency across time-steps.

PRAM-R is designed as an \textbf{asynchronous dual-loop architecture} that integrates a fast \textit{reactive loop} for real-time perception and control with a slow \textit{deliberative loop} for modality routing, reasoning, and memory updates. The reactive loop operates at high frequency to ensure immediate responsiveness, while the deliberative loop runs asynchronously at a lower frequency to refine sensing strategies based on accumulated context. Both loops involve memory interaction to maintain temporal consistency and adaptive behavior. 

At each reactive timestep $t$, the framework receives the current multimodal sensory streams 
$\mathcal{S}_t = \{\mathcal{S}_t^{\text{cam}}, \mathcal{S}_t^{\text{lidar}}, \mathcal{S}_t^{\text{radar}}\}$,
together with map data $\mathcal{M}_t^{\text{map}}$, prompts $\mathcal{P}_t$ and external information $\mathcal{E}_t$ (e.g., traffic signals, V2X messages, and weather conditions). 
It also accesses the latest available memory state $\mathcal{M}_{\tau}$, where $\tau \leq t$ denotes the most recent memory update produced by the deliberative loop. 
The reactive process generates the driving command $\mathcal{O}_t$ as:
\begin{equation}
    \mathcal{O}_t = f_{\text{react}}(\mathcal{S}_t, \mathcal{M}_t^{\text{map}}, \mathcal{E}_t, \mathcal{P}_t, \mathcal{M}_{\tau}),
\end{equation}
where $f_{\text{react}}(\cdot)$ represents the real-time perception–action mapping conditioned on current sensory inputs, environmental priors, and the latest memory snapshot.

Concurrently, the deliberative loop aggregates sensory, map, and action histories within a temporal window 
$\mathcal{H}_{t-\Delta\:t} = \{\mathcal{S}_{t-\Delta\:t}, \mathcal{M}_{t-\Delta\:t}^{\text{map}}, \mathcal{E}_{t-\Delta\:t}\}$ 
and updates the memory state via:
\begin{equation}
    \mathcal{M}_{t} = f_{\text{delib}}(\mathcal{H}_{t-\Delta\:t}, \mathcal{P}_{t-\Delta\:t}, \mathcal{M}_{t-\Delta\:t}),
\end{equation}
where $f_{\text{delib}}(\cdot)$ encapsulates LLM-guided modality routing, semantic abstraction, and long-term policy refinement.
This asynchronous dual-loop design allows PRAM-R to achieve both real-time responsiveness and high-level adaptivity under dynamic and uncertain driving conditions.

\subsection{Perception}

The perception module of PRAM-R integrates heterogeneous sensory inputs, including \textit{camera}, \textit{LiDAR}, and \textit{radar}, together with \textit{map} and \textit{external information}. There are three steps in the perception layer: shallow perception, modality routing and deep perception.

\begin{table*}[t]
\caption{Diagnostic and Contextual Indicators for Modality Routing}
\label{tab:indicators}
\centering
\renewcommand{\arraystretch}{1.15}
\setlength{\tabcolsep}{3pt}
\resizebox{\textwidth}{!}{%
\begin{tabular}{p{1.0cm}p{1cm}p{2.4cm}p{3.8cm}>{\centering\arraybackslash}p{0.9cm}p{2.6cm}p{3.8cm}p{1.1cm}}
\hline
\multicolumn{1}{c}{\raisebox{0pt}[2.5ex][0pt]{\textbf{Category}}} &
\multicolumn{1}{c}{\raisebox{0pt}[2.5ex][0pt]{\textbf{Modality}}} &
\multicolumn{1}{c}{\raisebox{0pt}[2.5ex][0pt]{\textbf{Diagnostic Indicators}}} &
\multicolumn{1}{c}{\raisebox{0pt}[2.5ex][0pt]{\textbf{Description}}} &
\multicolumn{1}{c}{\raisebox{0pt}[2.5ex][0pt]{\textbf{Enabled}}} &
\multicolumn{1}{c}{\raisebox{0pt}[2.5ex][0pt]{\textbf{Contextual Indicators}}} &
\multicolumn{1}{c}{\raisebox{0pt}[2.5ex][0pt]{\textbf{Interpretation}}} &
\multicolumn{1}{c}{\raisebox{0pt}[2.5ex][0pt]{\textbf{Reference}}} \\

\hline
\multirow[t]{12}{*}{Sensor} & \multirow[t]{3}{*}{Camera}
    & Brightness mean & Average gray-level intensity or exposure value within image patches. &
    \ding{51} & Illumination level &
    Low brightness → Low illumination → Low semantic segmentation accuracy. &
    \cite{sakaridis2019gcma} \\
  &  & Image contrast & Ratio of luminance difference between bright and dark regions. &
    \ding{51} & Visibility quality &
    Low contrast → Low visibility → Fog, rain, or night scenes. &
    \cite{ibrahim2019weathernet}\\
  &  & Edge density & Number of edges per image area, estimated by gradient magnitude. &
    \ding{51} & Texture density &
    Low edge density → Low texture density → Less structural information, degraded image clarity. &
    \cite{sakaridis2017sfsu}\\

  & \multirow[t]{5}{*}{LiDAR}
    & Point cloud density & Average number of LiDAR points per spatial cell. &
    \ding{51} & Lidar data completeness &
    Low density → Low data completeness → Sparse data. &
    \cite{heinzler2020cnn}, \cite{zhang2024image} \\
  &  & Noise ratio & Fraction of points with abnormal intensity or geometry. &
    \ding{51} & Lidar signal quality &
    High noise ratio → Low signal quality → fog or heavy rain. &
    \cite{heinzler2020cnn} \\
  &  & Reflectivity ratio & Ratio of measured to nominal reflectivity. &
    \ding{51} & Lidar signal quality &
    Low reflectivity ratio → Low signal quality → Fog or heavy rain. &
    \cite{bijelic2018lidar} \\
  &  & Echo ratio & Ratio between multiple returns and total echoes. &
    \ding{55} & N/A &
    N/A &
    \cite{man2021multi} \\
  &  & Z Variance & Variance of radar target heights within a temporal window. &
    \ding{55} & N/A &
    N/A &
    \cite{dong2024groundgrid} \\

  & \multirow[t]{4}{*}{Radar}
    & Target point density & Number of detected radar targets per frame or region. &
    \ding{51} & Radar data completeness &
    Low density → Low data completeness → Occlusion or signal degradation. &
    \cite{hunde2022multitarget}\\
  &  & Radar cross section & Magnitude of radar signal reflection from targets surface. &
    \ding{51} & Radar signal stability &
    Cross section fluctuations → Low signal stability → unstable detection, weather interference. &
    \cite{sevgi2013rcs}\\
  &  & Probability of detection hypothesis & Probability assigned to detected objects based on radar likelihood. &
    \ding{51} & Radar detection reliability &
    High probability → Low reliability → Unreliable perception. &
    \cite{caesar2020nuscenes}\\
  &  & Signal-to-noise ratio & Ratio of signal power to background noise. &
    \ding{55} & N/A &
    N/A &
    \cite{lukin2001noise} \\
\midrule
\multirow[t]{3}{*}{\makecell{External}}
  & \multirow[t]{3}{*}{N/A}
    & N/A & N/A &
    \ding{51} & Weather &
    Severe weather → degraded sensor reliability. &
    \cite{ibrahim2019weathernet} \\
  &  & Light & Global illumination intensity. &
    \ding{51} & Illumination level &
    Low illumination → Reduced visual clarity &
    \cite{ibrahim2019weathernet}  \\
\midrule
Map & N/A & N/A & N/A&
\ding{51} & Map scenes complexity &
High map complexity → rich contextual cues for reasoning. &
\cite{caesar2020nuscenes} \\
\bottomrule
\end{tabular}
}
\end{table*}

\subsubsection{Shallow Perception}

The perception layer performs \textit{shallow perception} to extract diagnostic indicators from multimodal sensors, the environment, and maps. Each indicator quantifies attributes such as reliability, completeness, and perception quality. As summarized in Table~\ref{tab:indicators}, these indicators are abstracted into higher-level \textit{contextual indicators} that guide reasoning-based modality routing and adaptive sensor selection. Camera indicators (brightness, contrast, edge density) describe illumination and texture; LiDAR indicators (point density, noise, reflectivity) assess signal quality; radar indicators capture detection reliability and stability. External cues like weather and map complexity provide complementary priors, while less informative or unavailable metrics (e.g., LiDAR echo ratio, radar Z variance, radar SNR) are omitted for efficiency. 

\subsubsection{Modality Routing}
Unlike conventional feature-level fusion, PRAM-R introduces an LLM-based \textbf{Qwen3-VL-8B} \textit{modality router} that performs LLM-guided routing directly at the perception level. 
At each timestep, the LLM-based router estimates both the scene-complexity–based usage mask and the reliability of each modality as: 
\begin{equation}
    \{u_i^t\}_{i=1}^{N},\ \{r_i^t\}_{i=1}^{N} 
    = f_{\text{LLM}}(\mathcal{I}_t, \mathcal{C}_t, \mathcal{M}_{t-1}),
\end{equation}
where $f_{\text{LLM}}(\cdot)$ denotes the Qwen3-VL-8B–based reasoning function that infers both the 
\textit{usage mask} $u_i^t\!\in\!\{0,1\}$ (reflecting scene complexity) and the 
\textit{reliability scores} $r_i^t\!\in\![0,1]$ from contextual indicators $\mathcal{I}_t$, 
scene context $\mathcal{C}_t$, and memory state $\mathcal{M}_{t-1}$.

The reliable modality set and corresponding confidence weights are obtained by threshold-based selection and proportional normalization: 
\begin{equation}
    \mathcal{R}_t = \{\, i \mid r_i^t \ge \theta_i \,\}, 
    \qquad
    w_i^t =
    \begin{cases}
        \dfrac{r_i^t}{\sum\limits_{j \in \mathcal{R}_t} r_j^t}, & i \in \mathcal{R}_t, \\[0.9em]
        0, & i \notin \mathcal{R}_t,
    \end{cases}
    \label{eq:weight_normalization}
\end{equation}

where $\theta_i$ denotes the reliability threshold for modality $i$. 
The normalized weights satisfy $\sum_i w_i^t = 1$, ensuring consistent contribution among the activated modalities while adaptively excluding unreliable ones. 
This proportional normalization provides stable and interpretable weighting for downstream sensor fusion.

In addition, the LLM also evaluates scene complexity to recommend a set of modalities that are sufficient for the current driving context:
\begin{equation}
    \mathcal{U}_t = \{\, i \mid u_i^t = 1 \,\},
\end{equation}
where $u_i^t \in \{0,1\}$ is the complexity-based usage indicator.  
The final activated modality set $\mathcal{F}_t$ is determined by combining both decisions:
\begin{equation}
\label{eq:intersection_fallback}
    \mathcal{F}_t = \mathcal{U}_t \cap \mathcal{R}_t,
    \qquad
    \text{if } |\mathcal{F}_t| = 0, \ \mathcal{F}_t \leftarrow \mathcal{R}_t.
\end{equation}
This intersection-then-fallback rule enables the router to 
activate only the minimal set of reliable modalities required for the current scene, 
thereby reducing redundant sensing and improving efficiency in simple environments,
while still maintaining robustness under complex or uncertain conditions.

To further stabilize routing decisions, hysteresis is applied to the same threshold used for modality selection. 
Let $\theta_i$ denote the reliability threshold for modality $i$, and define the on/off boundaries as 
$\theta_{\text{on}} = \theta_i + \delta$ and $\theta_{\text{off}} = \theta_i - \delta$. 
The activation state $s_i^t$ is then updated by:
\[
s_i^t =
\begin{cases}
1,& s_i^{t-1}=0\ \wedge\ r_i^t \ge \theta_{\text{on}},\\
0,& s_i^{t-1}=1\ \wedge\ r_i^t \le \theta_{\text{off}},\\
s_i^{t-1},& \text{otherwise.}
\end{cases}
\]
Continuous weights are further stabilized by an exponential moving average (EMA):
\begin{equation}
\tilde{w}_i^t =
\begin{cases}
w_i^0, & t = 0, \\[4pt]
\alpha w_i^t + (1-\alpha)\tilde{w}_i^{t-1}, & t > 0,
\end{cases}
\end{equation}
where $\tilde{w}_i^t$ denotes the temporally smoothed reliability weight used for sensor fusion, 
and $\alpha$ is the exponential smoothing factor controlling the trade-off between responsiveness and stability. 
The coefficient $\alpha$ is computed according to the frame interval $\Delta t$ and the desired smoothing time constant $\tau$ as $\alpha = 1 - e^{-\Delta t / \tau}$, where $\Delta t$ is the sampling period and $\tau$ determines how quickly the weights respond to changes. 

To implement the LLM-guided routing mechanism described above, Algorithm~\ref{alg:llm-modality-routing} summarizes the complete process of the LLM-based modality routing.

\begin{algorithm}[t]
\caption{LLM-based Modality Routing}
\label{alg:llm-modality-routing}
\KwIn{$\mathcal{I}_t$, $\mathcal{C}_t$, $\mathcal{M}_{t-1}$, $\{\theta_i\}_{i=1}^N$, $\delta$, $\Delta t$, $\tau$}
\KwOut{$\mathcal{F}_t$, $\{s_i^t\}$, $\{\tilde{w}_i^t\}$, $\mathcal{M}_t$}

$\{u_i^t\}, \{r_i^t\} \leftarrow f_{\mathrm{LLM}}(\mathcal{I}_t, \mathcal{C}_t, \mathcal{M}_{t-1})$\;

$\mathcal{U}_t \leftarrow \{i \mid u_i^t = 1\}$; 
$\mathcal{R}_t \leftarrow \{i \mid r_i^t \ge \theta_i\}$; 
$\mathcal{F}_t \leftarrow \mathcal{U}_t \cap \mathcal{R}_t$\;
\If{$|\mathcal{F}_t| = 0$}{$\mathcal{F}_t \leftarrow \mathcal{R}_t$}

$w_i^t \leftarrow r_i^t / \sum_{j \in \mathcal{F}_t} r_j^t$ for $i \in \mathcal{F}_t$; otherwise $w_i^t \leftarrow 0$\;

\For{$i \leftarrow 1$ \KwTo $N$}{
  $\theta_{\text{on}} \leftarrow \theta_i + \delta$; \quad $\theta_{\text{off}} \leftarrow \theta_i - \delta$\;
  \eIf{$s_i^{t-1}=0$ \textbf{and} $r_i^t \ge \theta_{\text{on}}$}{
    $s_i^t \leftarrow 1$\;
  }{
    \eIf{$s_i^{t-1}=1$ \textbf{and} $r_i^t \le \theta_{\text{off}}$}{
      $s_i^t \leftarrow 0$\;
    }{
      $s_i^t \leftarrow s_i^{t-1}$\;
    }
  }
}
$\alpha \leftarrow 1 - e^{-\Delta t / \tau}$; \quad
$\tilde{w}_i^t \leftarrow \alpha w_i^t + (1-\alpha)\tilde{w}_i^{t-1}$\;

$\mathcal{M}_t \leftarrow \textsc{UpdateMemory}(\mathcal{M}_{t-1}, \{u_i^t\}, \{r_i^t\}, \{s_i^t\}, \{\tilde{w}_i^t\})$\;
\Return{$\mathcal{F}_t, \{s_i^t\}, \{\tilde{w}_i^t\}, \mathcal{M}_t$}\;
\end{algorithm}

\subsubsection{Deep Perception}
After determining the selected modalities, the deep perception performs tasks such as object detection and semantic segmentation to extract information carried by each modality, including obstacle awareness and ego-vehicle state. The modality-specific features are aggregated through weighted fusion:
\begin{equation}
    \mathbf{F}_t^{\text{fused}} = \sum_{i \in \mathcal{R}_t} w_i\,\mathbf{F}_t^{(i)},
\end{equation}
where $w_i$ reflects the weight of modality $i$. An LLM-based (Qwen3-VL-8B) module further integrates the weighted fusion with multi-view camera inputs, providing high-level semantic understanding. 
The final perception output is stored as a structured JSON file that encapsulates semantic entities and environmental attributes. This unified, context-aware perception output is then transmitted to the reasoning layer for downstream scene analysis, decision making, and trajectory planning.

\subsection{Reasoning and Action}
The reasoning and action layers follow the design of PLA framework~\cite{zhang2025unified}. 
The reasoning layer employs Qwen3-VL-8B to transform fused perception into semantic representation, conducting scene-level analysis of spatial relationships and environmental cues. It produces driving intentions and risk-aware decisions, translated into trajectory plans and vehicle control commands in the action layer, ensuring safe navigation under dynamic conditions. Within PRAM-R, these layers are further coupled with the memory module to enable context-aware reasoning and policy refinement across timesteps.

\subsection{Memory Architecture and Update Mechanism}

\begin{table*}[t]
\centering
\caption{Trigger Conditions of Memory Components in the PRAM-R Framework}
\label{tab:memory_trigger}
\renewcommand{\arraystretch}{1.15}
\setlength{\tabcolsep}{3.5pt}
\begin{tabular}{p{2.6cm} p{3.5cm} p{3.5cm} p{3.5cm} p{3.5cm}}
\hline
\multicolumn{1}{c}{\textbf{Component}} &
\multicolumn{1}{c}{\textbf{Generation}} &
\multicolumn{1}{c}{\textbf{Reading}} &
\multicolumn{1}{c}{\textbf{Updating}} &
\multicolumn{1}{c}{\textbf{Expiration}} \\
\hline
\textit{P-Routing Context} 
& System start or first valid diagnostics 
& Before and after each routing decision 
& When new diagnostics or context arrive
& After next round of routing decision\\ 

\textit{P-Semantic Cache} 
& First deep-perception output in scene 
& Before each reasoning cycle 
& New deep-perception result
& After each reasoning cycle \\ 

\textit{R-Scene Record} 
& First completed reasoning cycle 
& Planning request or logging call 
& After every reasoning cycle 
& Scene end \\ 

\textit{R-Seed State} 
& After first reasoning cycle 
& At the start of the next reasoning cycle 
& When significant context change or periodic refresh 
& Scene/route reset\\ 

\textit{A-Policy Log} 
& First control command in scene 
& Validation or diagnostics 
& Each control cycle 
& Scene end \\ 

\textit{Knowledge Repository} 
& Distill from all other memory components with quality threshold met
& Scene reference or experience analysis
& Continuous consolidation
& Policy pruning or dataset management \\ 
\hline
\end{tabular}
\end{table*}

\subsubsection{Hierarchical Organization}
\begin{figure}[t]
    \centering
    \includegraphics[width=0.95\columnwidth]{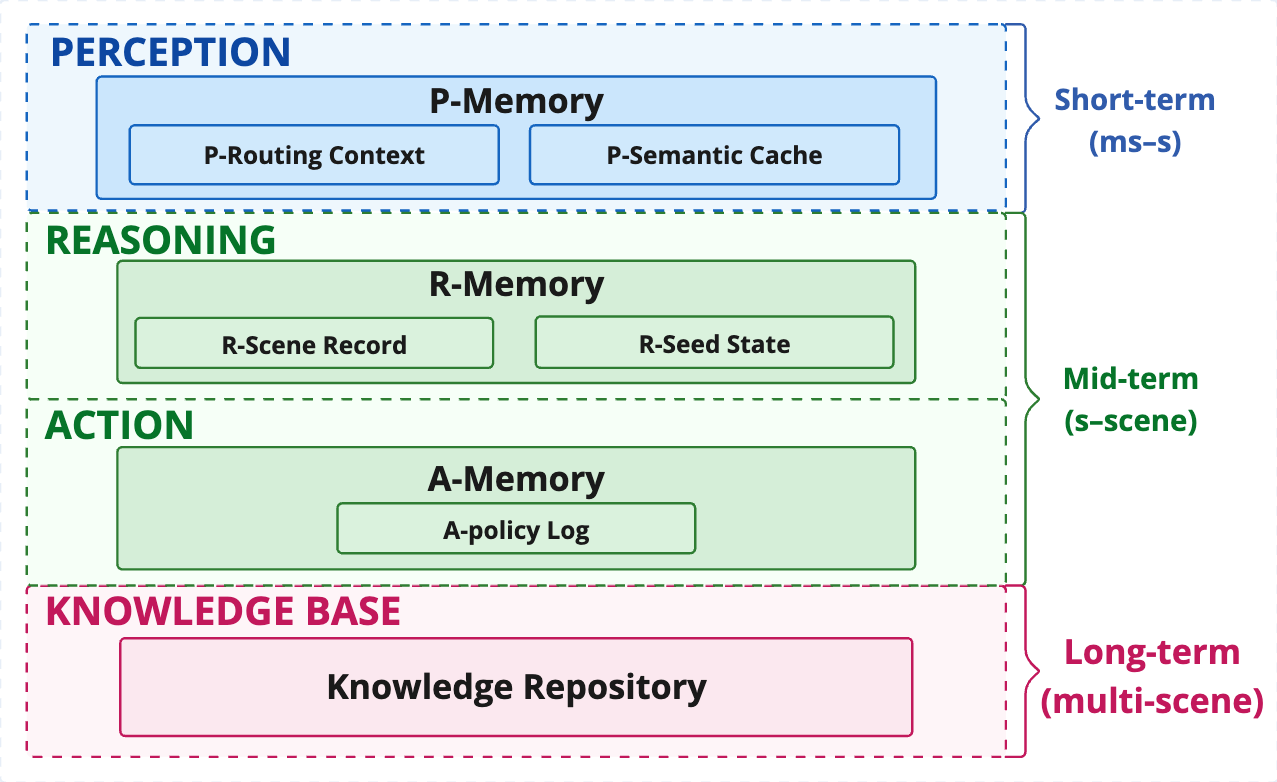}
    \caption{Hierarchical and sequential organization of PRAM-R memory components and their temporal scales. }
    \label{fig:memory_arch}
\end{figure}

To enable continuous reasoning and adaptation, PRAM-R employs four-layer hierarchical memory:\textit{perception}, \textit{reasoning}, \textit{action}, and a \textit{knowledge base}, each operating at a distinct temporal scale from short-term context to long-term experience. Fig.~\ref{fig:memory_arch} illustrates their sequential organization, enabling consistent perception, reasoning, and decision-making across timescales.

\textbf{a) Perception Module.} This module manages short-term perception states. 
The \textit{P-Routing Context} tracks diagnostic and contextual indicators along with routing decisions, while the \textit{P-Semantic Cache} compresses fused perception outputs—such as object detections, semantic segmentations, and map patches—into compact semantic representations for the reasoning layer.

\textbf{b) Reasoning Module.}
Mid-term reasoning memory maintains decision consistency and interpretability. The \textit{R-Scene Record} stores scene understanding and risk assessmentto guide reasoning in similar contexts, while the \textit{R-Seed State} retains abstract cues and policy hints for faster, semantically consistent inference.

\textbf{c) Action Module.}
At this module, the \textit{A-Policy Log} preserves execution continuity by recording control commands, trajectory errors, and constraint activations, linking high-level reasoning with low-level actuation to enable feedback-driven motion adaptation.

\textbf{d) Long-Term Knowledge Base.}
The \textit{Knowledge Repository} serves as long-term experiential memory, collecting distilled representations from perception, reasoning, and action after each episode. It stores high-level scene descriptors and optimal strategies, providing priors for future scenarios to enable continual self-improvement. 

\subsubsection{Operational Lifecycle of Memory Components}

The PRAM-R memory follows a cyclic, event-driven lifecycle with four phases: \textit{generation}, \textit{reading}, \textit{updating}, and \textit{expiration}.
Triggered by perception, reasoning, action, or scene transitions (Table~\ref{tab:memory_trigger}), short-term modules (\textit{P-Routing Context}, \textit{P-Semantic Cache}) synchronize with perception updates for rapid adaptation.
Mid-term modules (\textit{R-Scene Record}, \textit{R-Seed State}, \textit{A-Policy Log}) evolve with reasoning and control cycles to preserve temporal coherence, while the long-term \textit{Knowledge Repository} asynchronously consolidates experience for continual policy refinement.
This design achieves efficient memory management, balancing real-time operation with long-term stability and adaptability.

\section{Experiments}

To comprehensively validate the proposed \textit{PRAM-R} framework, we conduct a series of experiments on both synthetic and the real-world \textit{nuScenes} dataset. The evaluation focuses on routing adaptivity, stability, and efficiency under LLM-guided modality selection with hierarchical memory. 
\subsection{Experimental Setup}
\subsubsection{Baselines}
We compare PLA~\cite{zhang2025unified}, which activates all modalities without memory, and PRAM-R, which employs LLM-guided routing with hierarchical memory and hysteresis smoothing.

\begin{figure}[!t]
\centering
\begin{subfigure}[t]{0.24\textwidth}
    \centering
    \includegraphics[width=\linewidth]{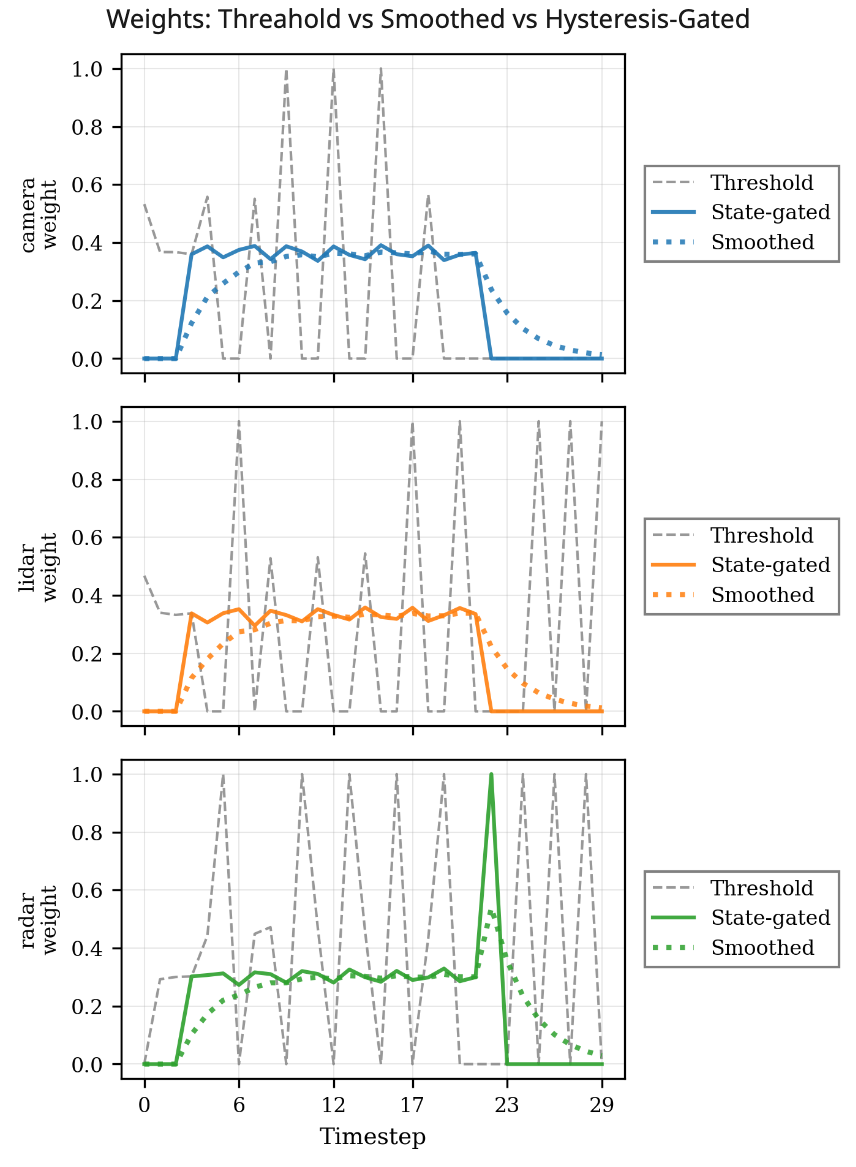}
    \caption{Weights}
    \label{fig:weights_hyst}
\end{subfigure}
\hfill
\begin{subfigure}[t]{0.24\textwidth}
    \centering
    \includegraphics[width=\linewidth]{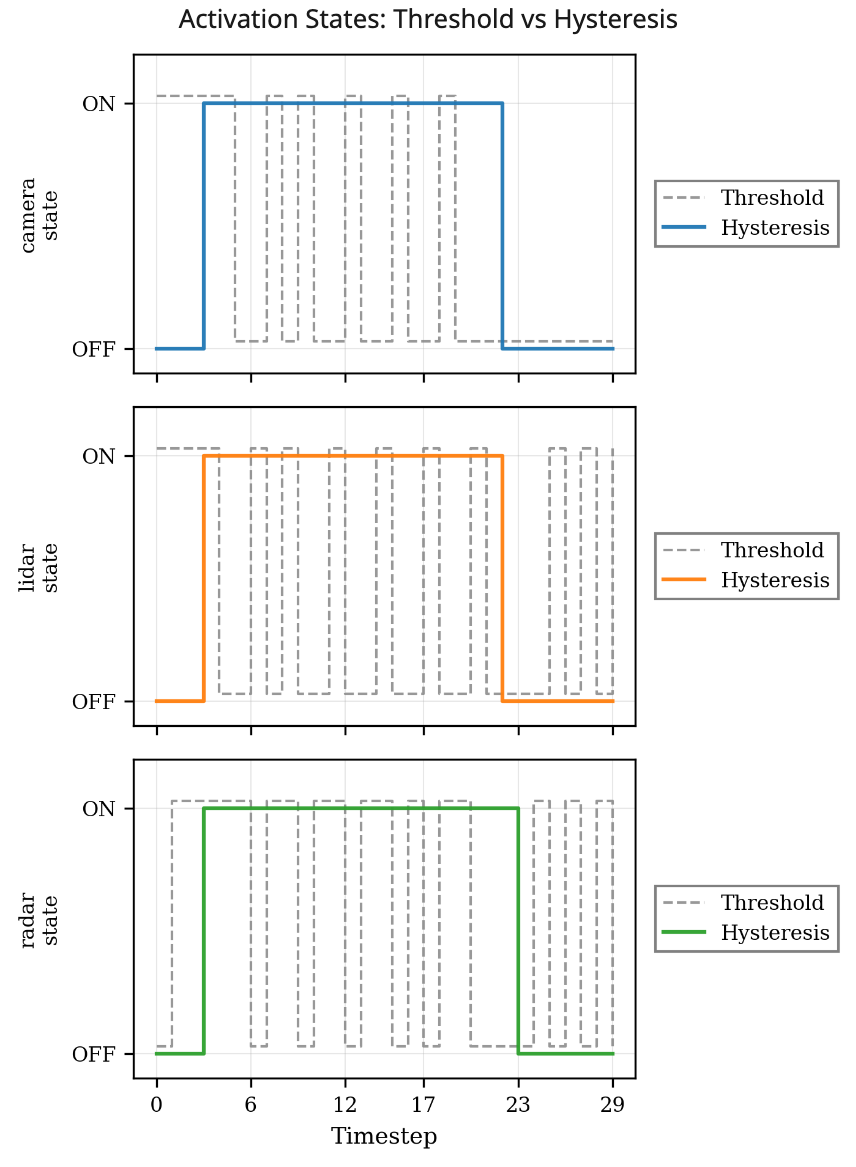}
    \caption{Activation States}
    \label{fig:states_hyst}
\end{subfigure}
\caption{\textbf{Threshold-Fluctuation Stress Test.}
Comparison of routing behavior under high-frequency perturbations.  (a) Weights before and after EMA and hysteresis gating across timestep.  (b) Corresponding binary activation states.}
\label{fig:threshold_stress_test}
\end{figure}

\subsubsection{Metrics}
For the assessment of routing adaptivity and stability, we employ the following functional and efficiency indicators: 

\textbf{a) Routing Efficiency (RE)}: ratio of deactivated modalities to total modalities, reflecting sensing load reduction and computational efficiency. 

\textbf{b) Routing Consistency (RC)}: temporal Jaccard similarity between consecutive routing masks, $ J(A,B) = \frac{|A \cap B|}{|A \cup B|} $, indicating decision smoothness.

\textbf{c) Routing Stability Index (RSI)}: normalized inverse standard deviation of modality weights, $ \mathrm{RSI} = 1 - \frac{\mathrm{std}(w_t)}{\bar{w}_t} $ representing the suppression of weight fluctuations through EMA and hysteresis filtering.

\textbf{d) Memory Recall Rate (MRR)}: percentage of successfully recalling previous reasoning results from memory. 

For evaluating perception and prediction performance, we adopt the average displacement error (ADE) and final displacement error (FDE)~\cite{alahi2016social} to measure trajectory accuracy and assess whether any performance degradation occurs.

\subsubsection{Prompt Specification}
\label{sec:prompt_spec}

The prompt instructs the LLM to perform sensor routing by mapping scene descriptions, contextual information, and numerical metrics to a strictly valid JSON output with predefined fields: 
(i) per-modality reliability scores in $[0,1]$ for camera, LiDAR, and radar,
(ii) a scalar scene complexity score in $[0,1]$, and
(iii) binary modality usage indicators for each modality.

\subsection{Routing Stability under Perturbation}
To evaluate routing stability under extreme conditions, we conduct synthetic stress tests with three perturbation types: gradual degradation simulating adverse weather, abrupt failures modeling sensor malfunctions, and high-frequency noise representing rapid fluctuations.
We measure switching frequency and weight variance to assess hysteresis-based stabilization. 
Table~\ref{tab:stability_hysteresis} and Fig.~\ref{fig:threshold_stress_test} jointly evaluate routing stability under pure thresholding and the proposed hysteresis mechanism. 
Hysteresis reduces modality switching by \textbf{87.2\%}, effectively suppressing high-frequency toggling by latching ON/OFF states for longer dwell times, while EMA smoothing further dampens in-state fluctuations without extra switches, jointly confirming consistent stability improvements and reduced weight variation across modalities.

\begin{table}[!t]
\centering
\caption{Effect of Hysteresis on Routing Stability under Fluctuations (threshold-only vs hysteresis)}
\label{tab:stability_hysteresis}
\renewcommand{\arraystretch}{1.0}
\setlength{\tabcolsep}{3pt}
\begin{tabular}{lccc}
\hline
\textbf{Modality} & \textbf{Switches (Thr)} & \textbf{Switches (Hyst)} & \textbf{Reduction (\%)} \\
\hline
Camera & 11 & 2 & 81.8\% \\
LiDAR  & 18 & 2 & 88.9\% \\
Radar  & 18 & 2 & 88.9\% \\
\textbf{Total} & \textbf{47} & \textbf{6} & \textbf{87.2\%} \\
\hline
\end{tabular}
\end{table}
\begin{figure}[!t]
\centering
\includegraphics[width=0.9\columnwidth]{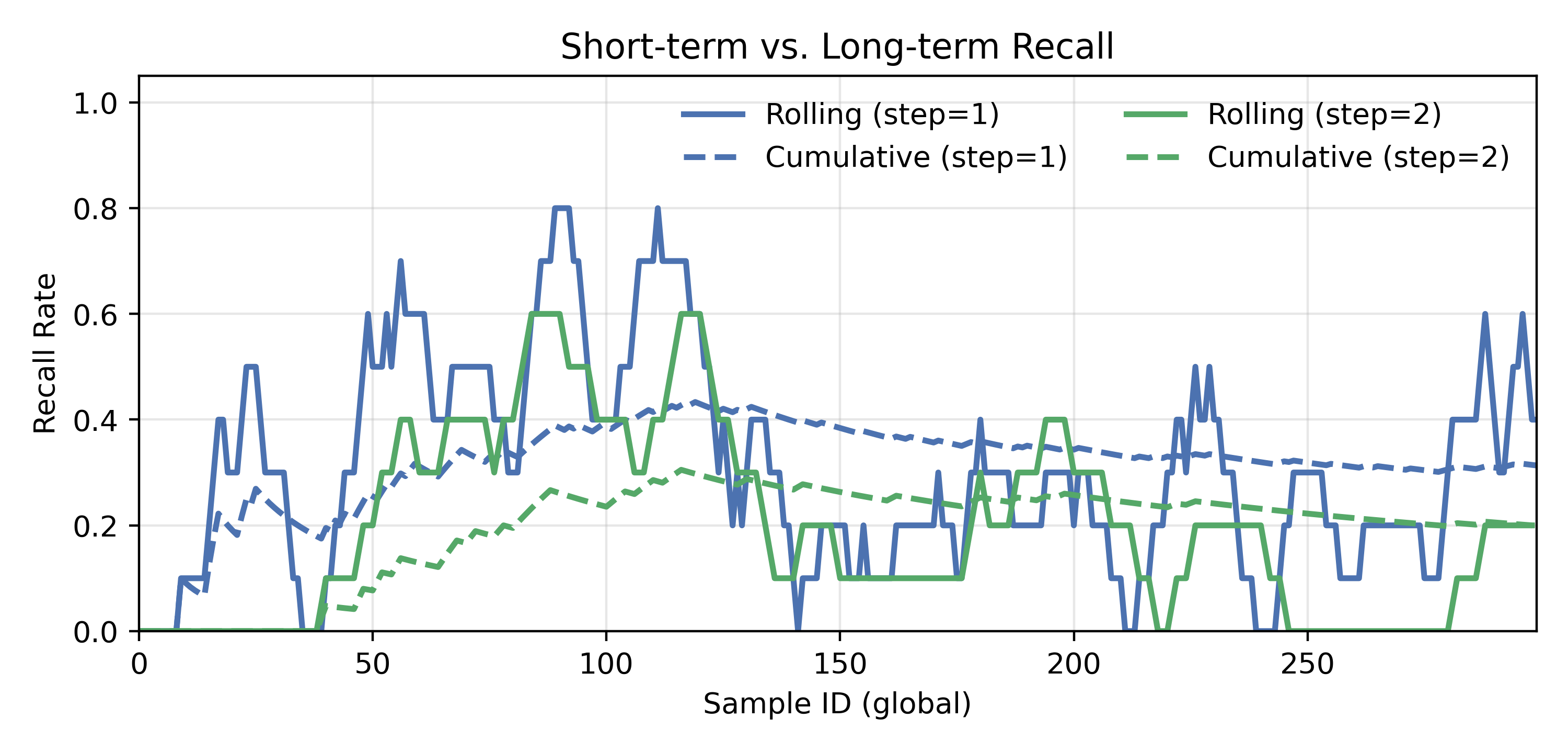}
\caption{\textbf{Combined Memory Recall Rate.}
Comparison of short-term (rolling) and long-term (cumulative) recall rates for two different sampling frequencies. 
}
\label{fig:recall_rate_combined}
\end{figure}

\subsection{Memory Recall Evaluation}

To analyze memory behavior, we visualize short- and long- term recall trends in Fig.~\ref{fig:recall_rate_combined}. Solid curves denote the rolling recall rate (short-term reuse within 5 seconds), while dashed curves show the cumulative recall ratio over time. Across different deliberative loop frequencies, PRAM-R maintains consistent reuse patterns: short-term recall adapts to scene transitions, and the long-term curve converges smoothly, indicating a balanced trade-off between adaptability and stability. 
And the effects of removing the hierarchical memory are discussed in the ablation study in \ref{Ablation}.

\begin{table}[!t]
\centering
\caption{Ablation study on memory and hierarchical routing components.}
\label{tab:ablation}
\renewcommand{\arraystretch}{1.0}
\setlength{\tabcolsep}{3pt}
\resizebox{\columnwidth}{!}{
\begin{tabular}{lcccccc}
\hline
\textbf{Variant} & \textbf{RE(\%)} & \textbf{RC} & \textbf{RSI} & \textbf{MRR} & \textbf{ADE(m)} & \textbf{FDE(m)} \\
\hline
Static Fusion (PLA) & 0.00 & 1.000 & 1.000 & 0.000 & 1.013 & 2.026 \\
PRAM-R (w/o Memory) & 6.02 & 0.992 & 0.662 & 0.000 & N/A & N/A \\
PRAM-R (w/o Dual-loop) & 5.67 & 0.998 & 0.641 & 0.313 & N/A & N/A \\
\textbf{Full PRAM-R} & \textbf{6.22} & \textbf{0.991} & \textbf{0.699} & \textbf{0.2000} & \textbf{1.124} & \textbf{2.182} \\
\hline
\end{tabular}}
\end{table}
\subsection{Overall Results and Ablation Study}
\label{Ablation}

\begin{figure*}[!t]
\centering
\begin{subfigure}[t]{0.29\textwidth}
    \centering
    \includegraphics[width=\linewidth]{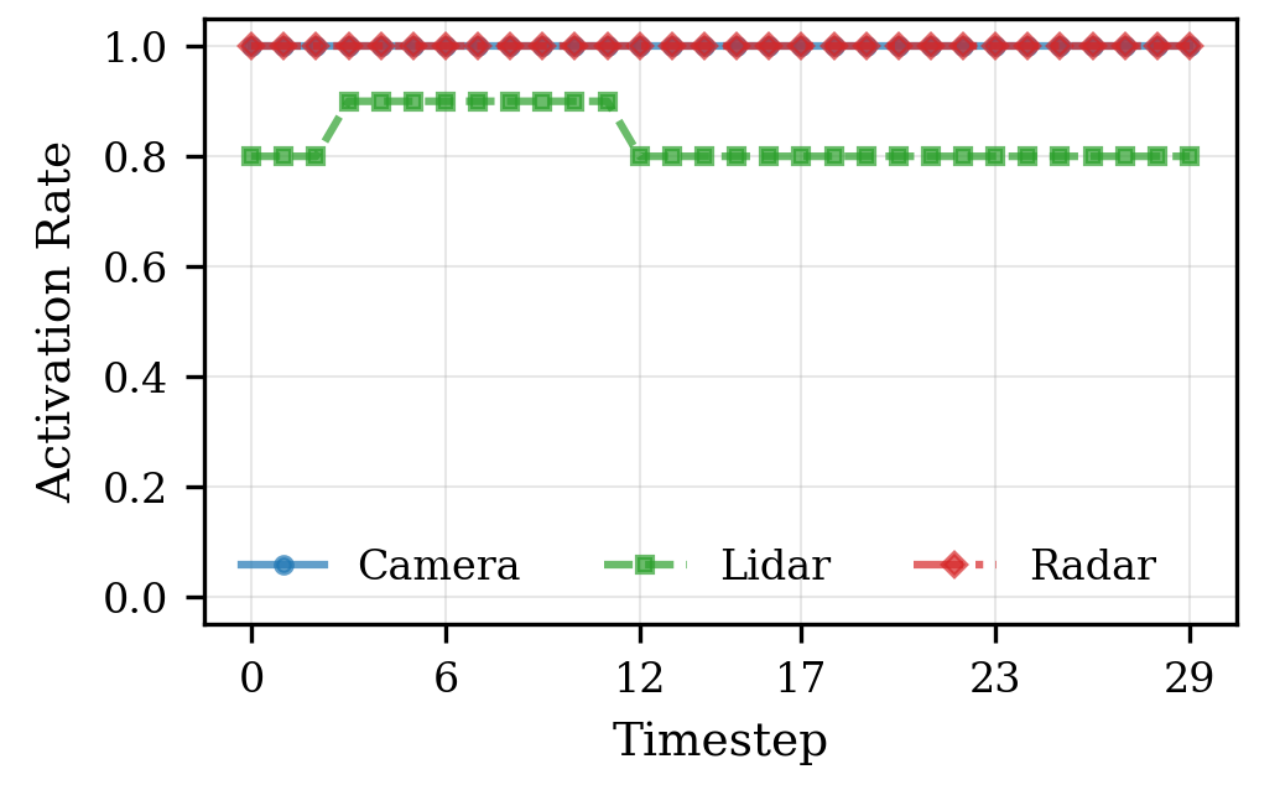}
    \caption{Activation Rate Over Time.}
    \label{fig:active_rate_curves}
\end{subfigure}
\hfill
\begin{subfigure}[t]{0.29\textwidth}
    \centering
    \includegraphics[width=\linewidth]{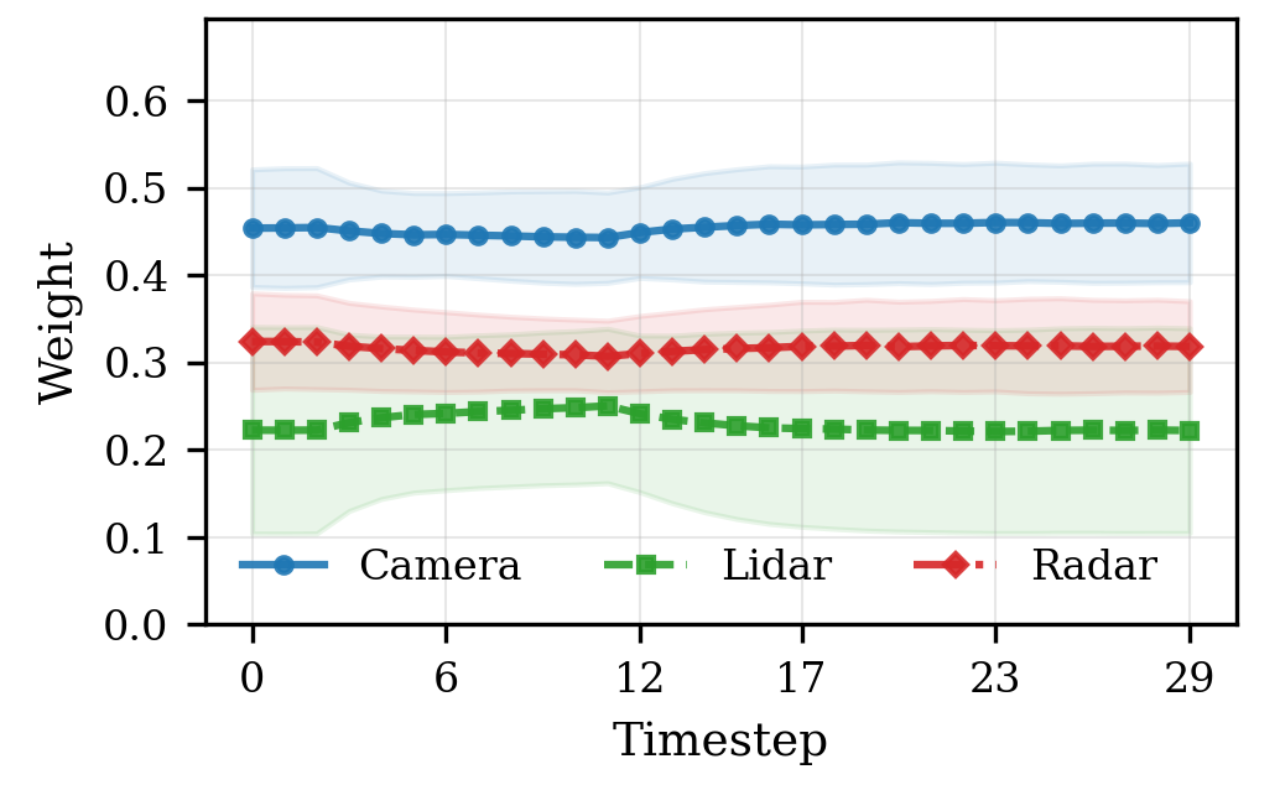}
    \caption{Mean Weights with Amplification Shading.}
    \label{fig:mean_weight_curves}
\end{subfigure}
\hfill
\begin{subfigure}[t]{0.28\textwidth}
    \centering
    \includegraphics[width=\linewidth]{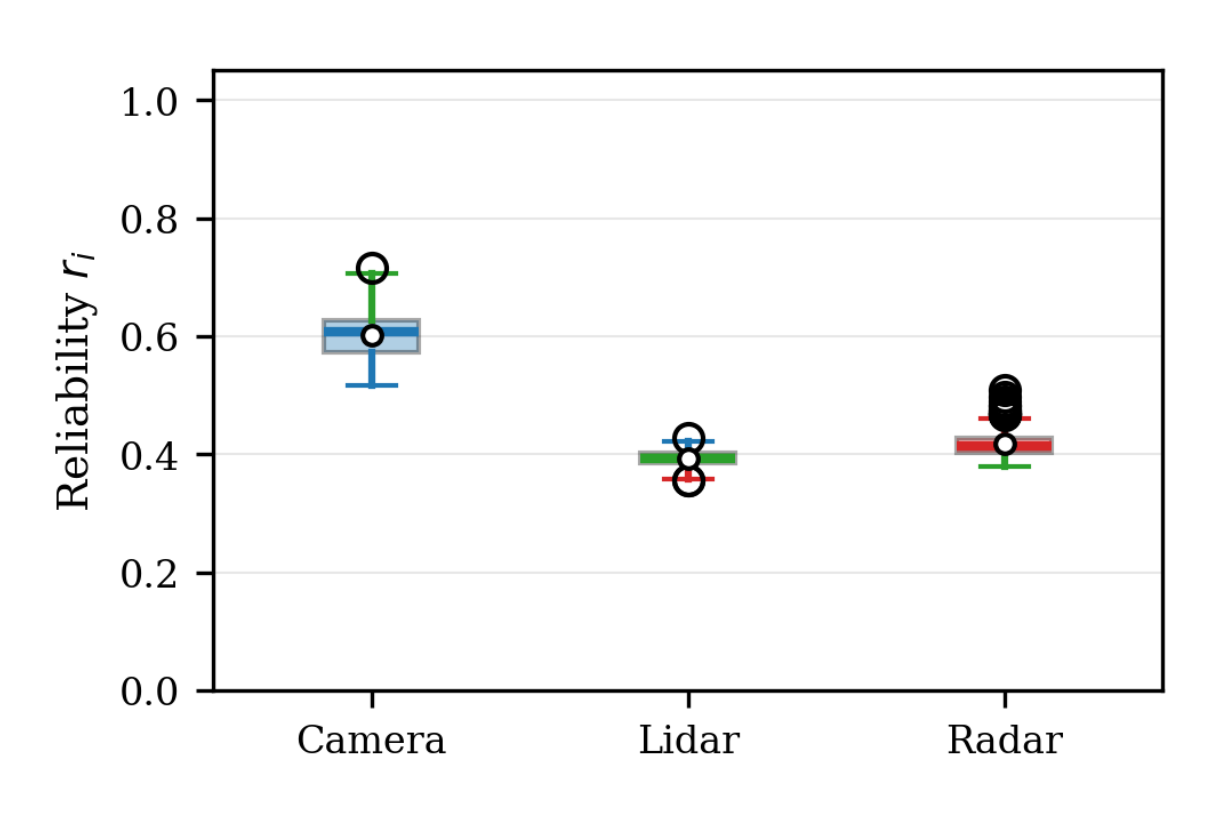}
    \caption{Reliability Distribution per Modality.}
    \label{fig:reliability_boxplot}
\end{subfigure}

\caption{\textbf{Comprehensive routing statistics across nuScenes.}
(a) Modality activation rates demonstrate adaptive engagement.
(b) Mean weight trajectories illustrate stable yet flexible fusion.
(c) Reliability distributions differs from each modality.}
\label{fig:dataset_level_eval}
\end{figure*}
We conduct an aggregated analysis on 10 representative scenes from the nuScenes dataset to assess activation rate, mean weight, and reliability distribution. 
Fig.~\ref{fig:active_rate_curves} shows that, in a specific setting of the reliability threshold, across the dataset, camera and radar provide continuous coverage (overlapped), while LiDAR adaptively deactivates in low-gain or noisy conditions, thereby reducing redundant computation. In practice, in medium-complex scenes, only camera and radar are active, whereas all sensors engage under high complexity. In Fig. ~\ref{fig:mean_weight_curves} and Fig. ~\ref{fig:reliability_boxplot}, the weight and reliability are consistent, without too much fluctuation. These global statistics complement the controlled stress tests, confirming that PRAM-R remains both adaptive and stable under diverse real-world driving scenarios.

As shown in Table~\ref{tab:ablation}, hierarchical memory and dual-loop reasoning consistently improve routing efficiency (RE) and stability (RSI) over the variants without memory or dual-loop (fast @2 Hz, slow @1 Hz) , with almost no loss in routing consistency (RC).
Compared with the static fusion baseline (PLA), PRAM-R reduces sensing and computational load with maximum routing efficiency (RE). Although memory recall rate (MRR) slightly decreases after applying dual-loop, this reflects stable activation and avoids redundant memory recalls. Perception metrics (ADE/FDE) remain comparable, confirming that PRAM-R achieves stable and efficient multimodal routing without degrading downstream performance.

\section{Conclusion and Future Work}
In summary, \textbf{PRAM-R} demonstrates the feasibility of combining LLM-guided routing and hierarchical memory for efficient multimodal perception and adaptive autonomous driving.
It achieves stable, context-aware sensor activation and efficient multimodal perception under diverse conditions. 
Experiments demonstrate significant gains in modality reduction and computational efficiency while maintaining prediction accuracy.

Despite promising results, several aspects remain for further exploration. 
\textbf{(1) Memory Validation:} due to space constraints, this work focuses on the core aspects of the memory mechanism; future work will include layer-wise memory ablations and quantification of memory’s impact on routing stability and downstream control.
\textbf{(2) Dataset and Deployment Gap:} small dataset and the absence of real-vehicle integration; future extensions will expand evaluation to real-vehicle logs and digital twins for broader validation. 
\textbf{(3) Model Size and Efficiency:} the current implementation uses Qwen3-VL-8B, introducing latency and memory overhead on embedded hardware; future work will quantify inference delay and efficiency to balance reasoning capability and real-time performance. 

\section*{Acknowledgment}
This research was funded by the Federal Ministry of Research, Technology and Space of Germany as part of the CeCaS project, FKZ: 16ME0800K.

\printbibliography

\end{document}